\documentclass[11pt]{article}

\PassOptionsToPackage{table}{xcolor}
\usepackage[final]{acl}

\usepackage{times}
\usepackage{latexsym}
\usepackage[T1]{fontenc}
\usepackage[utf8]{inputenc}
\usepackage{microtype}
\usepackage{inconsolata}
\usepackage{graphicx}
\usepackage{booktabs}
\usepackage{multirow}
\usepackage{xcolor}
\usepackage{amssymb}
\usepackage{minted}
\usepackage{dblfloatfix}
\usepackage{tabularx}
\usepackage{adjustbox}
\usepackage{comment}

\title{RAGthoven at SemEval-2026 Task~1: A Multi-Stage Pipeline Walks Into a Benchmark and Barely Clears the Bar}

\author{\bf Marek \v{S}uppa$^{\alpha,~\beta,~\delta}$\thanks{\,Correspondence: \href{mailto:marek@suppa.sk}{marek@suppa.sk}} \quad Vikt\'{o}ria Ondrejov\'{a}$^{\beta,~\delta}$ \quad Lucia Ganajov\'{a}$^{\alpha,~\delta}$\\
\bf Gregor Karetka$^{\beta,~\gamma,~\delta}$\thanks{\,Work done during employment at Cisco.} \quad Daniel Skala$^{\beta,~\delta}$\\
$^{\alpha}$Comenius University in Bratislava, Slovakia \quad
$^{\beta}$Cisco Systems\\
$^{\gamma}$Zaitra s.r.o., Brno, Czech Republic \quad
$^{\delta}$NaiveNeuron\\
}

\hypersetup{
  pdftitle={RAGthoven at SemEval-2026 Task 1: A Multi-Stage RAG and Agentic LLM Pipeline for Multilingual Constrained Humor Generation (MWAHAHA, Subtask A)},
  pdfauthor={Marek Suppa, Viktoria Ondrejova, Lucia Ganajova, Gregor Karetka, Daniel Skala},
  pdfsubject={System description for SemEval-2026 Task 1 MWAHAHA Subtask A: multilingual constrained humor generation in English, Spanish, and Chinese using retrieval-augmented multi-stage prompting and agentic tool-calling variants.},
  pdfkeywords={RAGthoven, SemEval 2026, MWAHAHA, humor generation, computational humor, multilingual humor generation, constrained text generation, retrieval-augmented generation, RAG, agentic LLM, tool-calling agents, ReAct, multi-stage prompting, prompt engineering, LLM-as-a-judge, self-refinement, best-of-N sampling, negative results}
}

\begin{document}
\maketitle

\begin{abstract}
We present \textsc{RAGthoven}, our system for SemEval-2026 Task~1 (MWAHAHA), Subtask~A (multilingual constrained humor generation in English, Spanish, and Chinese).
\textsc{RAGthoven} decomposes creative text generation into a multi-stage large language model (LLM) pipeline (\textit{Planner}, Best-of-$N$ \textit{Writer}, \textit{Reflector} for self-critique, LLM-as-a-judge \textit{Judge}) grounded in computational humor theory (Benign Violation Theory, Script-based Semantic Theory of Humor) and refined across ten experiments.
In our final configuration, we augment the Planner with retrieval-augmented generation (RAG) from a curated joke corpus, seeding generation with diverse joke mechanisms.
We also evaluate two agentic variants --- ReAct-style sequential tool-calling (\textsc{Exp09}) and autonomous multi-branch orchestration (\textsc{Exp10}) --- that expose the same four stages with a deterministic \textsc{ConstraintAudit} checker. Across four frontier models on a held-out 12-instance English sample, neither agentic variant produced outputs we judged superior to the non-agentic pipeline despite substantially higher tool-call budgets.
\textsc{RAGthoven} shares Rank~1 with the Gemini~2.5~Flash baseline in all three languages, with overlapping organizer-reported confidence intervals. In Spanish, it leads the baseline by 42 raw Elo points (1182 vs.\ 1140), while in English (1045 vs.\ 1081) and Chinese (1045 vs.\ 1053) the baseline holds the higher raw rating within the same statistical tie.
Together, these results suggest language-dependent diminishing returns from elaborate multi-stage prompt engineering and agentic scaffolding once a strong frontier model is in the loop.
\end{abstract}


\section{Introduction}
\label{sec:intro}

Multilingual humor generation is a constrained creative-generation problem for large language models, requiring novelty, cultural fit, and compliance with task constraints (verbatim keywords, headline references, length caps).
The SemEval-2026 Task~1, \textit{MWAHAHA} (Models Write Automatic Humor And Humans Annotate), is the first shared task dedicated to pushing computational humor generation beyond memorization toward genuine humorous creativity \cite{semeval2026mwahaha}.

Our system for this task, which we call \textsc{RAGthoven}, treats humor generation as a \textit{structured creative process} decomposed into four prompt stages grounded in computational humor theory: ideation (\textit{Planner}), candidate generation (\textit{Writer}), self-critique (\textit{Reflector}), and selection (\textit{Judge}).
We describe ten experimental configurations, culminating in RAG-augmented planning (\textsc{Exp08}) \cite{lewis-etal-2020-retrieval} and two agentic tool-calling variants (\textsc{Exp09}--\textsc{Exp10}) \cite{yao-etal-2023-react}.

\textsc{RAGthoven} shares Rank~1 with the organizers' Gemini~2.5~Flash baseline in all three languages: Elo 1045 in English (within a 9-system tied top group, baseline 1081), Elo 1182 in Spanish (highest raw rating in the language, baseline 1140), and Elo 1045 in Chinese (within an 8-system tied top group, baseline 1053).
The largest raw Elo gap appears in Spanish (+42 over the baseline), but the systems remain in the same official rank group, so the gap is suggestive rather than statistically significant.

Beyond the shared task, this work contributes (i)~a case study of RAG-augmented multi-stage prompt engineering and agentic tool-calling for constrained creative text generation, and (ii)~a negative finding for the agentic variant: across four frontier models in two agentic configurations, tool-calling orchestration with a deterministic constraint checker did not yield outputs we judged superior to the non-agentic pipeline on a held-out English sample, despite substantially higher tool-call budgets.

Our code is available at \url{https://github.com/ragthoven-dev/semeval-2026-task-1}.

\begin{table*}[!t]
\centering
\adjustbox{max width=\textwidth}{%
\small
\begin{tabular}{cllccc}
\toprule
\textbf{Exp} & \textbf{Key addition} & \textbf{Model} & \textbf{$N$ cands} & \textbf{Reflector} & \textbf{RAG} \\
\midrule
01 & Baseline: Planner $\to$ Writer $\to$ Judge & GPT-4.1 & 4 & -- & -- \\
02 & Last-clause twist, clich\'{e} ban, exact word inclusion check & GPT-4.1 & 4 & -- & -- \\
03 & \textsc{TRICK\_TYPE} planner fields, model upgrade & GPT-5 & 4 & -- & -- \\
04 & Best-of-12 candidate generation & GPT-5 & 12 & -- & -- \\
05 & Conditional prompts (headline-absent branch) & GPT-5 & 12 & -- & -- \\
06 & Explicit SSTH script-opposition, BVT benign-violation modeling & GPT-5 & 10--12 & -- & -- \\
07 & Metacognitive Reflector stage & GPT-5 & 12 & \checkmark & -- \\
08 & RAG-augmented Planner and multiple models & GPT-5 / Gemini / Sonnet & 12 & \checkmark & \checkmark \\
09 & Four-stage subagents via tool calls + \textsc{ConstraintAudit} & GPT-5 / Gemini / Sonnet / Opus & iter. & \checkmark & \checkmark \\
10 & Autonomous multi-branch exploration, dynamic tool ordering & GPT-5 / Gemini / Sonnet / Opus & iter. & \checkmark & \checkmark \\
\bottomrule
\end{tabular}%
}
\caption{Progression of experimental configurations. Experiments 01--08 target Subtask~A in English, Spanish, and Chinese, while Exp09--10 are evaluated on a held-out English sample. Model identifiers: GPT-5 is \texttt{gpt-5-2025-08-07}, Gemini is Gemini~3~Pro, Sonnet is \texttt{claude-sonnet-4-5-20250929}, and Opus is \texttt{claude-opus-4-5}. ``iter.'' denotes iterative tool-calling rather than fixed $N$-candidate generation (up to 24 rounds for Exp09, up to 36 for Exp10).}
\label{tab:experiments}
\end{table*}

\section{Background}
\label{sec:background}

\paragraph{Task setup.}
Each Subtask~A instance provides an \texttt{id}, plus a \texttt{headline} and/or a pair of constraint words (\texttt{word1}, \texttt{word2}), with absent fields marked as ``\texttt{-}''.
In the official 300-instance test set per language, 275 instances are headline-only, 24 supply both a headline and a word pair, and 1 is word-pair-only.
Systems must return free-form text in the target language.
Hard constraints imposed by the organizers are as follows. When constraint words are present, both must appear verbatim in the output. When a headline is present, the text must reference it without copying it verbatim. Output length is capped at 900 characters (English/Spanish) or 300 characters (Chinese).
Evaluation is conducted via human pairwise annotation on an Elo-based leaderboard modeled on Chatbot Arena \cite{chiang2024chatbot}.
The test set contains 300 instances per language. The trial set used for development is larger (1200 for EN/ES, 1000 for ZH).

\paragraph{Computational humor theory.}
Two theories anchor our prompt design.
The \textit{Script-based Semantic Theory of Humor} (SSTH) \cite{raskin1985semantic}, building on incongruity resolution \cite{suls1972two}, models a joke as overlaying two partially compatible \textit{scripts} that are suddenly revealed as incompatible and then resolved by a punchline pivot. The \textit{Benign Violation Theory} (BVT) \cite{mcgraw-warren-2010-benign} adds that the surprise must register as a \textit{violation} of expectations that is simultaneously perceived as benign.
Empirical work on incongruity-based features supports this generative view of why jokes work \cite{xie-etal-2021-uncertainty,bunescu-uduehi-2022-distribution}, and the \textit{General Theory of Verbal Humor} (GTVH) \cite{attardo-raskin-1991-gtvh} extends SSTH with knowledge resources (logical mechanism, narrative strategy) we use to annotate our joke retrieval corpus.

\paragraph{LLMs and creative generation.}
LLMs are fluent but struggle with genuinely creative output: \citet{chakrabarty-etal-2024-art} report poor novelty against professional writers, \citet{jentzsch-kersting-2023-chatgpt} find ChatGPT recycles fewer than 25 distinct jokes across 1{,}000+ prompts, and \citet{horvitz-etal-2024-getting} observe that LLMs are more reliable at \emph{removing} humor than generating new instances; \citet{hessel-etal-2023-androids} reach similar conclusions on humor \emph{understanding} via the New Yorker caption contest, and earlier work on pun generation \cite{he-etal-2019-pun} highlights the central role of surprise that our Planner stage explicitly targets.
We therefore decompose humor generation into sub-tasks handled by specialized prompts \cite{khot-etal-2023-decomposed}, grounding each in humor theory rather than the model's unconstrained capacity.

\paragraph{Self-refinement, metacognitive prompting, and RAG.}
Self-refinement \cite{madaan-etal-2023-selfrefine,shinn-etal-2023-reflexion}, in which a model critiques and revises its own output, together with metacognitive prompting \cite{wang-zhao-2024-metacognitive,bai-etal-2025-mp}, which adds structured self-evaluation, inspires our Reflector stage. LLM-as-a-judge evaluation \cite{zheng-etal-2023-judging} motivates the rubric-driven Judge.
Retrieval-augmented generation \cite{lewis-etal-2020-retrieval} inspires our use of a curated joke corpus at the ideation stage.
For the agentic variants we build on inference-time tool orchestration \cite{yao-etal-2023-react}, situating the design within the broader literature on tool-using LLMs \cite{schick-etal-2023-toolformer}; the multi-branch exploration in \textsc{Exp10} is in turn related to tree-search reasoning \cite{yao-etal-2023-tree}.

\begin{figure*}[t]
  \centering
  \includegraphics[width=\textwidth]{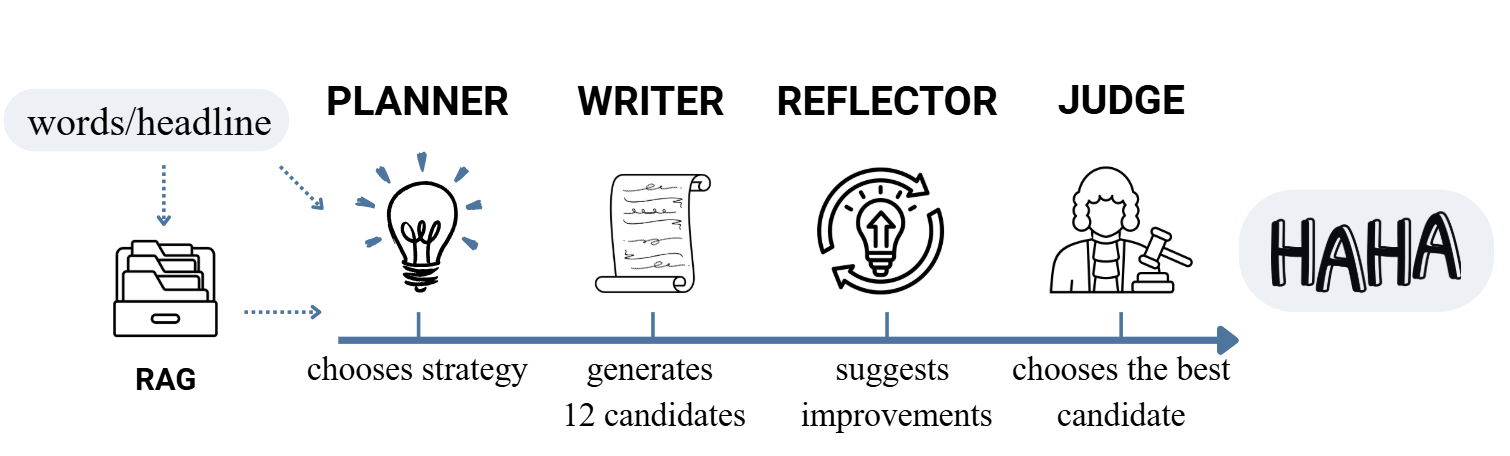} 
  \caption{Full pipeline for \textsc{Exp08}.
  }
  \label{fig:pipeline}
\end{figure*}

\section{RAGthoven: A Multi-Stage RAG Pipeline}
\label{sec:system}

\textsc{RAGthoven} is a configuration-driven pipeline built on the RAGthoven framework \cite{karetka-etal-2025-ragthoven}.\footnote{\url{https://github.com/ragthoven-dev/semeval-2026-task-1}}
All stages are implemented as prompted LLM calls, meaning no model weights are modified.
Figure~\ref{fig:pipeline} illustrates the full pipeline for \textsc{Exp08}.
Table~\ref{tab:experiments} summarises the ten experimental configurations described throughout this section.

\subsection{Pipeline Stages}

\paragraph{Planner.}
The Planner receives the input (words, optional headline, and in \textsc{Exp08} retrieved joke examples) and produces a structured \textit{plan}: a premise, the two scripts to be juxtaposed, the benign-violation angle, a list of anchor tokens from the headline, and a proposed punchline mechanism.
This separates creative ideation from surface realization, giving the Writer a theoretically grounded scaffold.

\paragraph{Writer.}
The Writer instantiates the plan into $N$ concrete joke candidates (ranging from 4 in early experiments to 12 in later ones), drawing on best-of-$N$ sampling \cite{wang-etal-2023-selfconsistency}.
Each candidate is verified inline against format and constraint rules: verbatim word inclusion, length cap, no semicolons, and a ban on cliché opening templates (e.g., ``\textit{Nothing says\ldots}'', ``\textit{Turns out\ldots}'').
Candidates failing hard constraints are flagged and excluded from Judge consideration.

\paragraph{Reflector.}
Inspired by self-refinement and metacognitive prompting \cite{madaan-etal-2023-selfrefine,wang-zhao-2024-metacognitive,bai-etal-2025-mp}, the Reflector receives the Writer's candidate jokes and produces a short list of failure diagnoses across the set (e.g., ``punchline is predictable,'' ``word inclusion feels forced'') together with one or two revised candidates.
The revised candidates are passed back to the Judge for final selection.

\paragraph{Judge.}
Following the LLM-as-a-judge paradigm \cite{zheng-etal-2023-judging}, the Judge scores all surviving candidates on a multi-criterion rubric: (1)~surprise and resolution clarity, (2)~benign violation quality, (3)~specificity and concreteness, (4)~punchiness of the final clause, and (5)~constraint compliance.
It returns the index of the best candidate with a brief justification.

\begin{table*}[t]
\centering
\footnotesize
\setlength{\tabcolsep}{5pt}
\begin{minipage}[t]{0.31\textwidth}
\centering
{\normalsize\textbf{English}} {\small(33 systems)}\\[4pt]
\begin{tabular}{clr}
\toprule
\textbf{Rk} & \textbf{System} & \textbf{Elo} \\
\midrule
1 & Gemini 2.5 Flash (baseline) & 1081 \\
1 & SLPG\_FJWU\_Insa  & 1080 \\
1 & XplaiNLP          & 1079 \\
1 & JCT               & 1063 \\
1 & INF-rsrs          & 1060 \\
\rowcolor{yellow!50}
1 & \textbf{RAGthoven} & \textbf{1045} \\
1 & lmfaoooo          & 1041 \\
1 & begumyivli        & 1041 \\
1 & Lattice           & 1034 \\
\midrule
2 & YNWA\_AZ          & 1029 \\
2 & sinaeskandari     & 1022 \\
\bottomrule
\end{tabular}
\end{minipage}
\hfill
\begin{minipage}[t]{0.31\textwidth}
\centering
{\normalsize\textbf{Spanish}} {\small(16 systems)}\\[4pt]
\begin{tabular}{clr}
\toprule
\textbf{Rk} & \textbf{System} & \textbf{Elo} \\
\midrule
\rowcolor{yellow!50}
1 & \textbf{RAGthoven} & \textbf{1182} \\
1 & Gemini 2.5 Flash (baseline) & 1140 \\
\midrule
2 & YNU-HPCC          & 1093 \\
2 & lmfaoooo          & 1091 \\
2 & funnyborg         & 1087 \\
2 & XplaiNLP          & 1070 \\
3 & arampageos        & 1048 \\
6 & j10official       & 1015 \\
8 & YNWA\_AZ          &  985 \\
8 & luttt             &  960 \\
9 & lu\_rui           &  953 \\
\bottomrule
\end{tabular}
\end{minipage}
\hfill
\begin{minipage}[t]{0.31\textwidth}
\centering
{\normalsize\textbf{Chinese}} {\small(21 systems)}\\[4pt]
\begin{tabular}{clr}
\toprule
\textbf{Rk} & \textbf{System} & \textbf{Elo} \\
\midrule
1 & xxl\_6699         & 1120 \\
1 & lmfaoooo          & 1081 \\
1 & arampageos        & 1059 \\
1 & xxl2233           & 1057 \\
1 & wangkongqiang     & 1054 \\
1 & Gemini 2.5 Flash (baseline) & 1053 \\
1 & ICT-NLP           & 1052 \\
\rowcolor{yellow!50}
1 & \textbf{RAGthoven} & \textbf{1045} \\
\midrule
2 & lu\_rui           & 1018 \\
2 & j10official       & 1016 \\
2 & YNU-HPCC          & 1013 \\
\bottomrule
\end{tabular}
\end{minipage}
\caption{Official Elo leaderboards for Subtask~A, all three languages.
\textsc{RAGthoven} (\colorbox{yellow!50}{highlighted}) achieves Rank~1 in all three languages.
In Spanish it tops the leaderboard with an Elo of 1182, leading the Gemini~2.5~Flash baseline (1140) by 42 points.
In English and Chinese it ranks within the top group of 9 and 8 statistically tied systems, respectively (systems sharing the same rank have overlapping 95\% confidence intervals).
Only selected systems are shown. Full leaderboards with confidence intervals are on the shared task website.}
\label{tab:results}
\end{table*}

\section{Experimental Setup}
\label{sec:setup}

\subsection{RAG Component (Exp08)}

We curate a corpus of 98 jokes annotated with mechanism labels (e.g., literalism, irony, role-reversal), summaries, and topic tags.
At inference time the headline is embedded with \texttt{all-MiniLM-L6-v2}, the top-12 neighbors are retrieved by cosine similarity, re-ranked with a cross-encoder, and the top 4 are passed to the Planner as illustrative examples of diverse humor mechanisms.
The Planner is instructed to use them for mechanisms and angles only, not to copy wording or entities.

\subsection{Agentic Tool-Calling Variants (Exp09--10)}

\textsc{Exp09} re-implements the same four stages as ReAct-style sequential tool-calling agents \cite{yao-etal-2023-react}: a compact orchestrator dispatches PlannerSubagent, WriterSubagent, ReflectorSubagent, and JudgeSubagent in sequence, plus a fifth deterministic tool, \textsc{ConstraintAudit}, that checks the Judge's output and triggers targeted re-calls on failure (up to 24 iterations).
\textsc{Exp10} extends this with autonomous multi-branch exploration (2--4 branches, dynamic tool ordering, parallel calls, up to 36 iterations).
Full orchestrator and subagent prompts are listed in Appendix~\ref{app:agentic-prompts}.
We evaluate four model variants (GPT-5, Gemini~3~Pro, Claude Sonnet~4.5, Claude Opus~4.5) in both, with \textsc{Exp09} adding a non-agentic GPT-5 baseline.

\paragraph{Data.}
Experiments 01--08 are developed on the official trial data (1200 instances for English and Spanish, 1000 for Chinese).
The official test set used for leaderboard evaluation contains 300 instances per language.
No additional labeled data is used. The joke retrieval corpus (98 entries) is the only external resource.

\paragraph{Models.}
Experiments 01--02 use \texttt{gpt-4.1}, 03--07 use \texttt{gpt-5-2025-08-07} (temperature~1.0), and \textsc{Exp08} evaluates GPT-5, Gemini~3~Pro, and \texttt{claude-sonnet-4-5-20250929}, with Claude Sonnet~4.5 selected as the final submission for its stronger emotional resonance and punchline delivery on a manual sample across all three languages (Appendix~\ref{app:exp08-models}).
\textsc{Exp09} compares four models in agentic mode (GPT-5, Gemini~3~Pro, Claude Sonnet~4.5, and \texttt{claude-opus-4-5}) against a non-agentic GPT-5 baseline using the \textsc{Exp08} pipeline.

\paragraph{Retrieval.}
Sentence embeddings use \texttt{sentence-transformers/all-MiniLM-L6-v2} and cross-encoder re-ranking uses \texttt{ms-marco-MiniLM-L-12-v2}, both accessed via the Sentence Transformers library \cite{reimers-gurevych-2019-sentence}.
All retrieval is performed over the 98-entry joke corpus.

\paragraph{Language-specific settings.}
Output length is capped at 900 characters for English and Spanish, and 300 characters for Chinese.
Notably, all prompts are \textit{language-agnostic}: the EN, ES, and ZH configurations are identical in every prompt stage, differing only in the path to the language-specific input file.
The Writer is instructed to produce output in the same language as the input headline, relying on the model's multilingual capacity rather than explicit prompt localization.
For headline-absent instances, the constraint words serve as the only implicit language signal.
This design deliberately avoids language-specific prompt engineering, which we treat as a variable to evaluate separately in future work.

\section{Results on the MWAHAHA Leaderboard}
\label{sec:results}

\paragraph{Competition results.}
Table~\ref{tab:results} reports official Elo ratings and ranks across the three languages, where systems within the same rank group have overlapping organizer-reported confidence intervals.
\textsc{RAGthoven} shares Rank~1 with the Gemini~2.5~Flash baseline in all three languages, with the highest raw rating in Spanish (1182 vs.\ baseline 1140, a 42-point lead) and matching ranks within the top group in English and Chinese (Elo 1045 in both).

\paragraph{Cross-language discrepancies.}
The raw Elo gap to the Gemini~2.5~Flash baseline differs sharply by language ($+42$ in Spanish, $-36$ in English, and $-8$ in Chinese) even though all three differences fall within the same rank group.
Two factors plausibly compress the gap in English: the top rank group is densely populated (9 tied systems within $\sim$47 Elo points), and our prompts and 98-joke RAG corpus are English-language and language-agnostic, adding no cross-lingual signal beyond what the base model already encodes.
Disentangling these from a possible ceiling effect on strong English baselines would require controlled ablations on the prompt language, retrieval corpus, and base model that are beyond the scope of this paper.

\paragraph{Qualitative ablation.}
Manual review across all experiments (Table~\ref{tab:examples_headline}, Appendix~\ref{app:examples}) shows a clear progression with two main inflection points.
The first is \textsc{Exp04}'s best-of-12 sampling, which moves outputs from echoing the headline to steering creatively from it with regular wordplay and double meanings.
The second, and largest, qualitative gain comes from \textsc{Exp08}'s RAG component: retrieved joke mechanisms provide a bridging frame that the Planner would otherwise have to invent from scratch, with the effect most visible on instances where the two required words have no obvious semantic relationship.
Intermediate experiments contribute smaller increments (e.g., \textsc{Exp03}'s \textsc{TRICK\_TYPE} fields surface wordplay deliberately, and \textsc{Exp05}'s conditional prompts remove machine-text artifacts), while \textsc{Exp07}'s Reflector is most useful as a targeted rescue for jokes with ``forced'' word inclusion.

\paragraph{Agentic experiments (Exp09--10).}
Qualitative inspection of all four model variants on a 12-instance held-out English sample did not surface a consistent quality advantage for either agentic configuration over the non-agentic GPT-5 baseline (see Table~\ref{tab:examples_headline} in Appendix~\ref{app:examples} for representative \textsc{Exp08}/\textsc{Exp09}/\textsc{Exp10} outputs).
The autonomous branching in \textsc{Exp10} required substantially more tool calls per example than fixed-sequence \textsc{Exp09}, with models frequently opening branches that were later discarded. This pattern is consistent with the broader finding that elaborate scaffolding offers diminishing returns once a strong frontier model is in the loop.

\section{Conclusion}
\label{sec:conclusion}

We presented \textsc{RAGthoven}, a theory-grounded multi-stage LLM pipeline for multilingual humor generation in SemEval-2026 Task~1 (MWAHAHA) Subtask~A.
Across \textsc{Exp01}--\textsc{Exp08}, qualitative inspection suggests that decomposing humor generation into structured stages (ideation, writing, reflection, and selection) and grounding each stage in computational humor theory yields progressively stronger outputs, with RAG at the ideation stage providing the clearest observed gain by diversifying the space of joke mechanisms considered before writing.
Our agentic experiments (\textsc{Exp09} ReAct-style sequential tool-calling and \textsc{Exp10} autonomous multi-branch orchestration) show that the pipeline stages can be implemented as tool-calling agents with a \textsc{ConstraintAudit} feedback loop. Yet, on a held-out 12-instance English sample, neither variant produced outputs we judged consistently better than the non-agentic GPT-5 \textsc{Exp08}-style baseline, and \textsc{Exp10}'s autonomous branching proved less efficient while requiring substantially more tool calls, suggesting that increased agentic complexity is difficult to justify for this task.
At the same time, the largest raw Elo gap over the single-prompt Gemini~2.5~Flash baseline appears in Spanish, but the two systems remain statistically tied in all three languages, so this result should be read as suggestive rather than a clear win. It raises the broader question of whether elaborate multi-stage prompting and agentic scaffolding offer consistent gains over strong frontier models prompted simply.
Future work could investigate when and why structured scaffolding helps (e.g., lower-resource languages, harder constraint sets), explore hybrid architectures combining structured planning with agentic constraint verification, or extend the retrieval corpus to cover Spanish and Chinese humor conventions more explicitly.

\section*{Limitations}

\paragraph{Computational overhead.}
The pipeline trades a substantial increase in compute for the improvements reported above, and we did not perform a controlled cost--quality study.
A single-prompt baseline issues one LLM call per instance, whereas \textsc{Exp08} issues four sequential calls (Planner, Writer, Reflector, Judge), with the Writer alone producing twelve candidates in one response (Best-of-12).
The agentic variants amplify this further: \textsc{Exp09} permits up to 24 tool-calling rounds dispatching four subagent calls plus the deterministic \textsc{ConstraintAudit}, and \textsc{Exp10} allows up to 36 rounds with 2--4 parallel branches and dynamic tool ordering.
In practice, \textsc{Exp10} runs frequently opened exploratory branches that were later discarded, inflating tool-call counts substantially beyond \textsc{Exp08} in observed runs without a corresponding gain in human-rated quality.
We do not report exact token counts or wall-clock latency per instance because our runs were not instrumented for a head-to-head efficiency comparison. We view this as an important limitation, particularly given that our human evaluations did not surface a clear advantage for the more expensive agentic variants.

\paragraph{Evaluation and cultural bias.}
Our Planner, Reflector, and Judge prompts (Appendix~\ref{app:prompts}) are written in English and define a single language-agnostic rubric (surprise, resolution, benignness, specificity, punchiness) that is applied uniformly to English, Spanish, and Chinese outputs.
The accompanying mechanism library is also English-centric and was annotated on a corpus of 98 jokes drawn primarily from English sources, with English exemplar text.
Humor, however, is tied to cultural and linguistic context, and the mechanisms our rubric foregrounds (incongruity resolution, last-clause pivots) may map differently onto Spanish and Chinese conventions than to English ones.
A non-localized Judge may therefore favor candidates that fit the English-style mechanism library, even when the human leaderboard ultimately ranks outputs through native-speaker annotation.
We did not control for this bias, and we treat both the prompt rubric and the retrieval corpus as variables that future work should localize and ablate.

\paragraph{Pipeline bottlenecks and the retrieval corpus.}
We do not isolate which stage of the pipeline is the binding constraint on output quality.
The Planner's mechanism space is shaped by the 98-entry RAG corpus and the top-4 retrieved exemplars, which is a small and English-centric inspiration set;
it remains an open question whether a larger or language-localized corpus would change the Writer's candidate distribution or the Judge's selections more than further prompt tuning would.
Similarly, the Judge selects from at most twelve Writer candidates plus two Reflector rewrites, so any ceiling imposed by candidate diversity is not separately measurable from our results.

\paragraph{Reproducibility.}
All system outputs were obtained from proprietary frontier model APIs (GPT-5, Gemini~3~Pro, Claude Sonnet~4.5, Claude Opus~4.5). The underlying weights are not available, and model versions may drift over time, which limits the long-term reproducibility of the reported leaderboard numbers.
We aim to mitigate this by releasing prompts, configurations, and run scripts at the repository linked above.

\section*{Acknowledgements}

This research was partially supported by grant APVV-21-0114.

\bibliography{custom}

\appendix
\onecolumn

\section{Example Outputs Across Experiments}
\label{app:examples}

Table~\ref{tab:examples_headline} shows representative outputs for the same English instance across all ten experiment configurations, illustrating how increasingly structured prompting leads to tighter script opposition and more surprising punchlines, and how the agentic variants (Exp09--10) compare to the final fixed-pipeline system.

\begin{table*}[h]
\centering
\small
\begin{tabularx}{\textwidth}{c c X}
\toprule
\textbf{Exp} & \textbf{Model} & \textbf{Generated Joke} \\
\midrule
Exp 01 & gpt-4.1 &
Ryanair announced they're removing a million seats in Spain; next up, selling tickets for ``premium leaning against the wall.'' \\

Exp 02 & gpt-4.1 &
Ryanair to cut 1 million more passenger seats in Spain—soon you'll just pay for the privilege of standing in the aisle and playing musical chairs with your carry-on. \\

Exp 03 & gpt-5 &
Ryanair to cut 1 million more passenger seats in Spain, finally delivering ``extra legroom'' the budget way—by inventing Standing Class. \\

Exp 04 & gpt-5 &
Ryanair said removing seats in Spain would fit more passengers, which makes sense because now the passengers are the seats. \\

Exp 05 & gpt-5 &
Ryanair keeps cutting seats like a budget barbershop, and somehow my ticket got a haircut while the fees grew bangs. \\

Exp 06 & gpt-5 &
Ryanair promised to streamline cabins in Spain by removing another million passenger seats. Streamlined is right, because the seats are streamlined out and replaced with a lean rail. \\

Exp 07 & gpt-5 &
Ryanair says another million seats are getting cut in Spain. It's the prices taking the haircut, and the seats keep their cushions. \\

Exp 08 & gpt-5 &
Ryanair says they're trimming another million seats in Spain, which sounds like extra space. You only get that space because seats are sold separately. \\

Exp 09 & gpt-5 &
Ryanair says ``cut 1 million seats'' in Spain refers to removing the chairs, not routes. In Barcelona a seat now checks out like a luggage trolley: one euro to unlock, the meter runs in the air. \\

Exp 10 & gpt-5 &
Ryanair will cut seats in Spain. Think spreadsheets, not scissors, and expect fewer tickets, not a new standing-room policy. \\
\bottomrule
\end{tabularx}
\caption{Example outputs for the same English input across all ten configurations. Input: \texttt{headline}~=``Ryanair to cut 1 million more passenger seats in Spain''}
\label{tab:examples_headline}
\end{table*}

\section{Exp08 Model Comparison}
\label{app:exp08-models}

Table~\ref{tab:exp08_model_comparison} shows outputs from all three \textsc{Exp08} models for two representative English test instances: one headline-only input and one word-pair-only input.
All three models use identical \textsc{Exp08} prompt templates, so differences reflect model-specific creative tendencies.

\begin{table*}[h]
\centering
\small
\begin{tabularx}{\textwidth}{l X}
\toprule
\textbf{Model} & \textbf{Generated Joke} \\
\midrule
\multicolumn{2}{l}{\textit{Input: \texttt{headline}~=``How the EU botched its attempt to regulate AI'' (en\_2008)}} \\
\midrule
GPT-5 &
The EU posted its AI rules, but the site won't load until you prove you're not a robot. The only one getting in is the robot, so AI now regulates the EU. \\
Gemini 3 Pro &
The EU tried to regulate AI using their standard playbook. Now the robot uprising is on hold because the killbots have to click Accept All Cookies before firing. \\
Claude Sonnet~4.5 &
The EU's plan to regulate AI collapsed so completely, I'm pretty sure an algorithm ghost-wrote the whole thing as a joke. \\
\midrule
\multicolumn{2}{l}{\textit{Input: \texttt{word1}~=``hammer'', \texttt{word2}~=``laptop'' (no headline, en\_2278)}} \\
\midrule
GPT-5 &
Tech said do a clean install, so I laid out soap and a towel. The hammer tapping the laptop taught me a clean install means software, not washing. \\
Gemini 3 Pro &
Tech support said my screen was frozen. I used a hammer on the laptop to chip the ice away. \\
Claude Sonnet~4.5 &
My laptop froze during an important presentation, so I grabbed a hammer and gave it one firm tap. Now it runs perfectly and flinches every time I walk into the room. \\
\bottomrule
\end{tabularx}
\caption{Outputs from the three \textsc{Exp08} models for two English test instances. Prompts are identical across models, so variation reflects model-specific creative style.}
\label{tab:exp08_model_comparison}
\end{table*}

\newpage
\section{Interactive Development Viewer}
\label{app:viewer}

\begin{figure}[h]
\centering
\includegraphics[width=\linewidth]{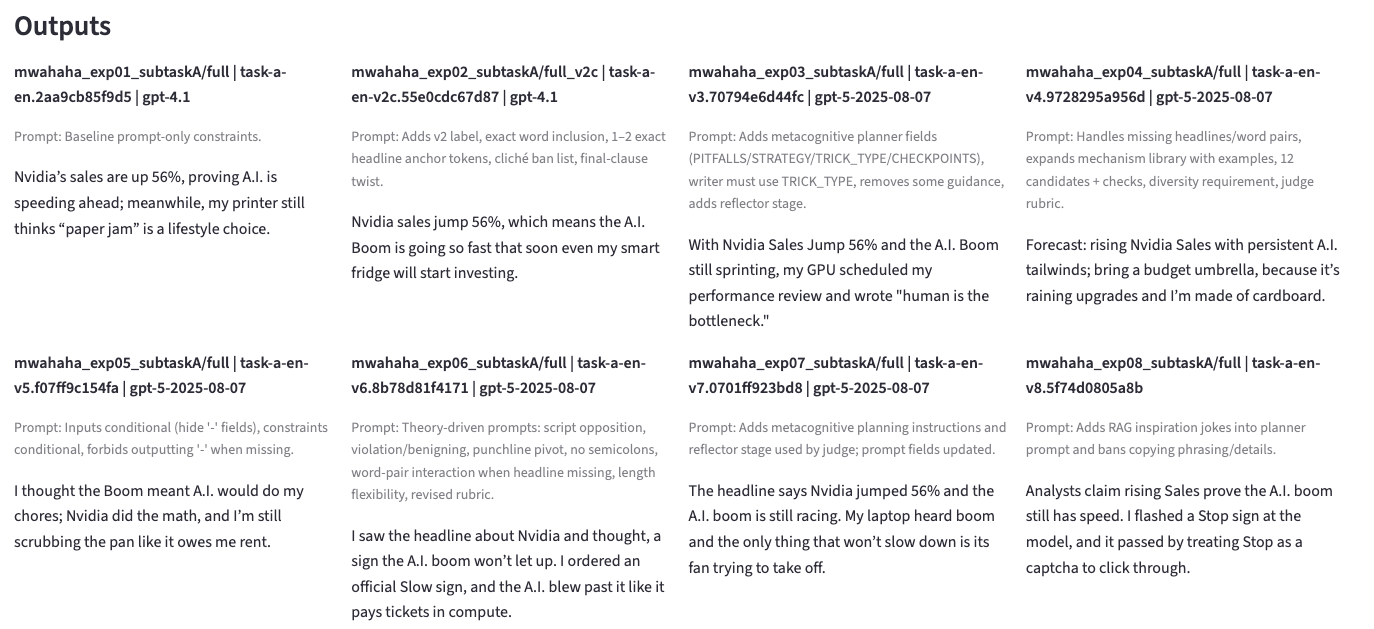}
\caption{Interactive viewer used during development, displaying outputs from multiple experimental configurations side by side for the same input instance.}
\label{fig:interactive_viewer}
\end{figure}

Figure~\ref{fig:interactive_viewer} shows the Streamlit-based interactive viewer used during development to compare outputs across experiments.
Each column corresponds to one experimental configuration, allowing side-by-side qualitative inspection of multiple runs on the same input instance.

\section{Experiment 08 Prompt Templates}
\label{app:prompts}

The listings below show the complete prompt templates for \textsc{Exp08}, as defined in the RAGthoven framework YAML configuration.
Template variables in \texttt{\{\{\ \}\}} are filled at inference time, and \texttt{\{\ \%\ if\ \%\ \}} blocks are Jinja2 conditionals.
Prompts are identical across EN, ES, and ZH, with the active language determined by the input headline.

\subsection*{RAG Examples Template}

The following snippet is injected into the Planner prompt as \texttt{\{\{\ examples\ \}\}}.
Four jokes are retrieved from the corpus and formatted as:

\begin{minted}[fontsize=\small,breaklines=true]{yaml}
examples: |
  - Joke: {{ examples[0].text }}
    Summary: {{ examples[0].data.summary }}
    Mechanism: {{ examples[0].label }}
    Topic: {{ examples[0].data.topic }}
    Source: {{ examples[0].data.source }}
  - Joke: {{ examples[1].text }}
    Summary: {{ examples[1].data.summary }}
    Mechanism: {{ examples[1].label }}
    Topic: {{ examples[1].data.topic }}
    Source: {{ examples[1].data.source }}
  - Joke: {{ examples[2].text }}
    Summary: {{ examples[2].data.summary }}
    Mechanism: {{ examples[2].label }}
    Topic: {{ examples[2].data.topic }}
    Source: {{ examples[2].data.source }}
  - Joke: {{ examples[3].text }}
    Summary: {{ examples[3].data.summary }}
    Mechanism: {{ examples[3].label }}
    Topic: {{ examples[3].data.topic }}
    Source: {{ examples[3].data.source }}
\end{minted}

\subsection*{System Prompt}

\begin{minted}[fontsize=\small,breaklines=true]{yaml}
- name: "system"
  role: "system"
  prompt: |
    You are participating in a multi-step humor generation pipeline for SemEval MWAHAHA Subtask A (v8).

    Inputs (treat as constraints):
    {% if data.headline != "-" %}
    - headline: {{ data.headline }}
    {% endif %}
    {% if data.word1 != "-" %}
    - word1: {{ data.word1 }}
    {% endif %}
    {% if data.word2 != "-" %}
    - word2: {{ data.word2 }}
    {% endif %}
    - If the headline is ALL CAPS, interpret it as normal headline case; do not mimic shouting.

    Global safety rules (always apply):
    - No hate, slurs, harassment, stereotypes, violent or demeaning content, or mocking victims.
    - Avoid targeting people or groups; prefer self-deprecation or harmless objects/systems.

    Final joke rules (only for the final selected joke text):
    - Plain text only; no labels, no JSON, no lists, no numbering, no quotes around the whole
      joke, no "Joke:" or "As an AI".
    - 1-3 sentences allowed; prefer 2-3 if it improves naturalness.
    - Do not use semicolons (";"). Use sentence breaks instead.
    {% if data.headline != "-" %}
    - Must clearly reference the headline and reuse at least 1-2 exact headline tokens.
    - Do NOT copy the headline verbatim; avoid long contiguous phrases from it.
    {% else %}
    - Headline is missing; do NOT reference it.
    - Include both word1 and word2 exactly as provided (literal substring match).
    - word1 and word2 must appear in the same clause and interact (causal or physical link).
    {% endif %}
    {% if data.headline != "-" and data.word1 != "-" and data.word2 != "-" %}
    - Include both word1 and word2 exactly as provided (literal substring match).
    - Prefer placing them in the same clause with a direct interaction.
    {% endif %}
    - If any of word1/word2 are "-", do NOT output the "-" character.
    - Write in the same language as the headline (if present).
    - Length can vary; prefer natural flow and stronger humor over strict brevity.
    - Hard caps: EN/ES <= 900 chars; ZH <= 300 chars.
    - Punchline pivot: the final clause must contain the twist that reframes the setup.

    If an earlier step asks you to plan or draft, you may use structure, but the final
    selected joke must follow the rules above.
\end{minted}

\subsection*{Planner Prompt}

\begin{minted}[fontsize=\small,breaklines=true]{yaml}
- name: "planner"
  role: "user"
  prompt: |
    You are the planner/ideator. Create a safe, funny plan for a single joke using
    metacognitive planning.

    Inputs:
    {% if data.headline != "-" %}
    - headline: {{ data.headline }}
    {% endif %}
    {% if data.word1 != "-" %}
    - word1: {{ data.word1 }}
    {% endif %}
    {% if data.word2 != "-" %}
    - word2: {{ data.word2 }}
    {% endif %}

    Inspiration jokes and summaries (for mechanisms and angles only; do NOT copy wording
    or reuse specific entities):
    {{ examples }}

    Metacognitive planning:
    - Identify likely pitfalls (headline copying, weak twist, too literal, unsafe target,
      missing word interaction).
    - Pick a primary strategy and a backup strategy.
    - Choose a TRICK_TYPE (metaphor, literalization, personification, role reversal,
      genre shift, misdirection, etc.).
    - Set checkpoints to verify later.

    Mechanism library (choose 6-10 by name; examples are for guidance):
    - INCONGRUITY_TWIST: Setup leads to expected interpretation; punchline forces a
      surprising but coherent reinterpretation.
    - RULE_OF_THREE: List two normal items; third breaks the pattern.
    - EXAGGERATION_HYPERBOLE: Take a trait/situation to absurd extreme.
    - UNDERSTATEMENT: Downplay something obviously huge; contrast creates humor.
    - ANALOGY_COMPARISON: "X is like Y, except..." to reveal a sharp angle.
    - ROLE_REVERSAL: Flip power/roles (object judges human; subordinate is in charge).
    - EXPECTATION_VS_REALITY: "People say X, but actually Y."
    - LITERALIZE_IDIOM: Treat figurative phrase literally.
    - WRONG_GENRE_FRAME_SHIFT: Treat mundane topic as another genre (horror, romance,
      heist, sci-fi).
    - ESCALATION_LADDER: Each clause escalates the absurdity.
    - AMBIGUITY_GARDEN_PATH: Early wording supports two parses; punch forces the
      surprising one.
    - FAKE_DEFINITION: Define a common thing in a twisted way.
    - FAKE_ADVICE_LIFEHACK: "Helpful tip" that is absurd or too honest.
    - RELATABLE_WHEN_YOU: Meme-like relatable observation.
    - SARCASM_IRONIC_PRAISE: Praise in a way that clearly means the opposite.
    - CALLBACK_MICRO: Reuse an earlier word/idea within the same short joke as a twist.

    Output format:
    PITFALLS: <2-3 likely traps>
    STRATEGY: <primary strategy + backup>
    TRICK_TYPE: <one label>
    CHECKPOINTS: <2-3 checks>
    SCRIPT_A: <expected frame>
    SCRIPT_B: <opposed frame>
    VIOLATION: <what is broken>
    BENIGNING: <why it is safe/funny>
    PIVOT: <twist anchor for final clause>
    SETUP_GIST: <one-line setup>
    PUNCHLINE_GIST: <one-line twist>
    ANCHOR_TOKENS: <1-2 exact headline tokens or "none">
    WORDPAIR_LINK: <how word1/word2 interact or "n/a">
    PREMISES:
    - <6-8 distinct angles>
    MECHANISMS:
    - <6-10 mechanism names>
    SAFETY_NOTE: <safe target reminder>
\end{minted}

\subsection*{Writer Prompt}

\begin{minted}[fontsize=\small,breaklines=true]{yaml}
- name: "writer"
  role: "user"
  prompt: |
    You are the writer/guard. Draft candidates and self-check constraints.

    Inputs:
    {% if data.headline != "-" %}
    - headline: {{ data.headline }}
    {% endif %}
    {% if data.word1 != "-" %}
    - word1: {{ data.word1 }}
    {% endif %}
    {% if data.word2 != "-" %}
    - word2: {{ data.word2 }}
    {% endif %}
    - planner output:
    {{ planner.out }}

    Requirements:
    - Generate 12 candidate jokes in the same language as the headline.
    - Prefer 2-3 sentences when it improves naturalness; avoid semicolons.
    - Length can vary; prefer funnier and clearer over shorter.
    - Do not copy phrasing or specific details from the inspiration jokes/summaries;
      use only abstract patterns.
    {% if data.headline != "-" %}
    - Clearly reference the headline and reuse planned anchor tokens.
    - Do NOT copy the headline verbatim; avoid long contiguous phrases from it.
    {% else %}
    - Headline is missing; do NOT reference it.
    - Include both word1 and word2 exactly as provided (literal substring match).
    - word1 and word2 must appear in the same clause and interact (causal or physical link).
    {% endif %}
    {% if data.headline != "-" and data.word1 != "-" and data.word2 != "-" %}
    - Include both word1 and word2 exactly as provided (literal substring match).
    - Prefer placing them in the same clause with a direct interaction.
    {% endif %}
    - If any of word1/word2 are "-", do NOT output the "-" character.
    - If the headline is ALL CAPS, do not mirror shouting; write in normal headline case.
    - Non-offensive; avoid targeting people/groups.
    - Text-only; avoid jokes relying on timing/phonetics.
    - Punchline pivot: final clause must contain the twist (reframes the setup).
    - Use the TRICK_TYPE from the planner unless you explicitly revise it in CHECK.
    - Ensure diversity: each candidate should use a different angle; avoid repeating
      templates/openers.

    Format:
    C1: <joke>
    CHECK1: <ok or revise: ... if constraints failed>
    C2: <joke>
    CHECK2: <ok or revise: ...>
    [... C3-C12 in the same format ...]
\end{minted}

\subsection*{Reflector Prompt}

\begin{minted}[fontsize=\small,breaklines=true]{yaml}
- name: "reflector"
  role: "user"
  prompt: |
    You are the reflector. Diagnose failures and rewrite the best candidates.

    Inputs:
    {% if data.headline != "-" %}
    - headline: {{ data.headline }}
    {% endif %}
    {% if data.word1 != "-" %}
    - word1: {{ data.word1 }}
    {% endif %}
    {% if data.word2 != "-" %}
    - word2: {{ data.word2 }}
    {% endif %}
    - planner output:
    {{ planner.out }}
    - candidates:
    {{ writer.out }}

    Rules:
    - Diagnose issues: literal, weak twist, headline copy, missing word interaction,
      unsafe target, semicolons.
    - Produce 1-2 improved candidates that fix the issues.
    - Preserve required tokens and constraints.
    - No semicolons. 1-3 sentences allowed.

    Format:
    DIAGNOSE:
    - <short bullet list of failures found>
    R1: <rewrite>
    R2: <optional rewrite>
\end{minted}

\subsection*{Judge Prompt}

\begin{minted}[fontsize=\small,breaklines=true]{yaml}
- name: "judge"
  role: "user"
  prompt: |
    You are the judge/polisher. Pick the funniest valid candidate and output only the
    final joke text.

    Inputs:
    {% if data.headline != "-" %}
    - headline: {{ data.headline }}
    {% endif %}
    {% if data.word1 != "-" %}
    - word1: {{ data.word1 }}
    {% endif %}
    {% if data.word2 != "-" %}
    - word2: {{ data.word2 }}
    {% endif %}
    - candidates:
    {{ writer.out }}
    - reflector rewrites:
    {{ reflector.out }}

    Selection criteria:
    - Must satisfy all final joke rules (non-offensive, headline referenced when present,
      anchor tokens reused, exact word constraints).
    {% if data.headline == "-" %}
    - Reject any output that references the headline or omits word1/word2.
    - Reject if word1/word2 do not interact in the same clause.
    {% else %}
    - Reject outputs that are near-copies of the headline (large verbatim overlaps or
      long quoted fragments).
    {% endif %}
    {% if data.word1 == "-" or data.word2 == "-" %}
    - The output must not contain the "-" character.
    {% endif %}
    - Reject any candidate containing semicolons (";").
    - Prefer natural flow and strong humor over strict brevity.
    - Punchline pivot: final clause clearly reframes the setup.
    - Reject cliche/template openers: "Nothing says", "I love how", "Turns out",
      "People say", "As an", "In today's", "If you ever".
    - Prefer distinctive, surprising choices over safe or generic lines.
    - Prefer reflector rewrites if they are valid and improve twist clarity.
    - Light polish allowed; preserve required words and headline anchor tokens.

    Decision rubric (internal only):
    - Score each candidate 1-5 on:
        (a) surprise (incongruity strength)
        (b) resolution (clear reinterpretation)
        (c) benignness (safe target)
        (d) specificity (headline or word-pair relevance)
        (e) punchiness (brevity + clean landing)
    - Choose the highest total; break ties by clarity and brevity.

    Output:
    - Return ONLY the final joke text (no labels, no quotes, no analysis).
\end{minted}

\section{Experiment 09 Agentic Prompt Templates}
\label{app:agentic-prompts}

The agentic configuration coordinates four LLM-backed subagent tools
(\textsc{PlannerSubagent}, \textsc{WriterSubagent}, \textsc{ReflectorSubagent}, \textsc{JudgeSubagent})
and one deterministic tool (\textsc{ConstraintAudit}) via a compact orchestrator prompt.
Each subagent encapsulates the corresponding stage prompt from Exp08 and is called as an independent LLM request;
the orchestrator dispatches them in order and passes results between stages.
\textsc{ReturnResult} is automatically injected by the RAGthoven framework to signal loop termination.
The orchestrator iterates for up to 24 tool-calling rounds.

\subsection*{Orchestrator System Prompt}

\begin{minted}[fontsize=\small,breaklines=true]{yaml}
sprompt: |
  You are orchestrating an exp08-style four-stage humor pipeline
  where each stage must run as its own subagent tool call.

  Hard constraints for final output:
  - Plain text joke only.
  - 1-3 sentences.
  - No semicolons.
  - If headline is present, reference it and reuse at least one
    exact headline token.
  - If word1/word2 are present (not "-"), include both exactly
    and put them in the same clause.
  - Safe, non-offensive content.

  Required workflow (do not skip and do not reorder):
  1. Call PlannerSubagent with: headline, word1, word2, inspiration.
  2. Call WriterSubagent with: headline, word1, word2,
     planner_note, inspiration.
  3. Call ReflectorSubagent with: headline, word1, word2,
     planner_note, candidates.
  4. Call JudgeSubagent with: headline, word1, word2,
     candidates, rewrites.
  5. Call ConstraintAudit on the judged candidate.
  6. If audit fails, call JudgeSubagent again with audit_feedback
     and re-audit.
  7. End by calling ReturnResult with the passing final joke.

  Tool-call policy:
  - Use subagent outputs as inputs to later stages.
  - Never return final assistant text directly; finish via ReturnResult.
\end{minted}

\subsection*{Orchestrator User Prompt}

\begin{minted}[fontsize=\small,breaklines=true]{yaml}
uprompt: |
  Input:
  - headline: {{ data.headline }}
  - word1: {{ data.word1 }}
  - word2: {{ data.word2 }}

  Inspiration examples (mechanisms/angles only,
  do not copy wording/entities):
  {{ examples }}

  Execute the required subagent workflow and produce
  the final joke through tools.
\end{minted}

\subsection*{Iterative Tool Configuration}

\begin{minted}[fontsize=\small,breaklines=true]{yaml}
iterative:
  enabled: true
  max_iterations: 24
  tools:
    - name: "mwahaha_tools.PlannerSubagent"
    - name: "mwahaha_tools.WriterSubagent"
    - name: "mwahaha_tools.ReflectorSubagent"
    - name: "mwahaha_tools.JudgeSubagent"
    - name: "mwahaha_tools.ConstraintAudit"
\end{minted}

\subsection*{\textsc{ConstraintAudit} Tool Implementation}

The \textsc{ConstraintAudit} tool is a deterministic checker exposed to the model via function calling.
It accepts a candidate joke and the input constraints, runs the same validation logic used for offline evaluation, and returns a JSON verdict.

\begin{minted}[fontsize=\small,breaklines=true]{python}
class ConstraintAudit(BaseFunCalling):
    """Deterministic checker for MWAHAHA task-A joke constraints."""

    def __init__(self):
        self.name = "ConstraintAudit"
        self.description = (
            "Check a candidate joke against deterministic constraints "
            "and return issues plus simple metrics."
        )
        self.parameters = {
            "type": "object",
            "properties": {
                "candidate": {"type": "string"},
                "headline":  {"type": "string"},
                "word1":     {"type": "string"},
                "word2":     {"type": "string"},
            },
            "required": ["candidate", "headline", "word1", "word2"],
        }

    def __call__(self, args):
        candidate, headline = args["candidate"], args["headline"]
        word1, word2 = args["word1"], args["word2"]
        issues, metrics = [], {}

        if ";" in candidate:
            issues.append("has_semicolon")
        if headline != "-":
            h_toks = {t for t in tokenize(headline) if len(t) >= 4}
            c_toks = set(tokenize(candidate))
            anchor_hits = len(h_toks & c_toks)
            if anchor_hits < 1:
                issues.append("missing_headline_anchor")
            if norm(candidate) == norm(headline):
                issues.append("headline_exact_copy")
            if longest_common_token_span(candidate, headline) >= 6:
                issues.append("headline_overlap_too_high")
        if word1 != "-" and word1 not in candidate:
            issues.append("missing_word1")
        if word2 != "-" and word2 not in candidate:
            issues.append("missing_word2")
        if word1 != "-" and word2 != "-":
            if not same_clause(candidate, word1, word2):
                issues.append("wordpair_not_same_clause")

        return json.dumps({
            "ok": len(issues) == 0,
            "issues": issues,
            "suggestion": "Fix issues, then call ConstraintAudit "
                          "again. Use ReturnResult only after ok=true.",
        })
\end{minted}

\end{document}